\documentclass[pdflatex,sn-basic]{sn-jnl}% Math 
\pdfoutput=1
\usepackage{graphicx}%
\usepackage{multirow}%
\usepackage{amsmath,amssymb,amsfonts}%
\usepackage{amsthm}%
\usepackage{mathrsfs}%
\usepackage[title]{appendix}%
\usepackage{xcolor}%
\usepackage{textcomp}%
\usepackage{manyfoot}%
\usepackage{booktabs}%
\usepackage{algorithm}%
\usepackage{algorithmicx}%
\usepackage{algpseudocode}%
\usepackage{listings}%
\usepackage{geometry}%

\usepackage{ulem}
\usepackage[compatibility=false]{caption}
\usepackage{adjustbox}
\input epsf
\usepackage{subfigure}
\usepackage{setspace}
\usepackage{float}
\usepackage{colortbl}
\usepackage{hyperref}

\theoremstyle{thmstyleone}%

\theoremstyle{thmstyletwo}%

\theoremstyle{thmstylethree}%

\begin{document}

\title[Article Title]{FastSmoothSAM: A Fast Smooth Method For Segment Anything Model}


\author[1]{\fnm{Jiasheng } \sur{Xu}}\email{xjs@stu.hqu.edu.cn}

\author*[1,2]{\fnm{Yewang} \sur{Chen}}\email{ywchen@hqu.edu.cn}

\affil*[1]{\orgdiv{The College of Computer Science and Technology}, \orgname{Huaqiao University}, \orgaddress{\city{Xiamen}, \postcode{361021}, \state{Fujian}, \country{China}}}

\affil[2]{\orgdiv{Key Laboratory of Computer Vision and Machine Learning (Huaqiao University)}, \orgaddress{\city{Xiamen}, \postcode{361021}, \state{Fujian}, \country{China}}}


\abstract{Accurately identifying and representing object edges is a challenging task in computer vision and image processing. The Segment Anything Model (SAM) has significantly influenced the field of image segmentation, but suffers from high memory consumption and long inference times, limiting its efficiency in real-time applications. To address these limitations, Fast Segment Anything (FastSAM) was proposed, achieving real-time segmentation. However, FastSAM often generates jagged edges that deviate from the true object shapes. Therefore, this paper introduces a novel refinement approach using B-Spline curve fitting techniques to enhance the edge quality in FastSAM. Leveraging the robust shape control and flexible geometric construction of B-Splines, a four-stage refining process involving two rounds of curve fitting is employed to effectively smooth jagged edges. This approach significantly improves the visual quality and analytical accuracy of object edges without compromising critical geometric information. The proposed method improves the practical utility of FastSAM by improving segmentation accuracy while maintaining real-time processing capabilities. This advancement unlocks greater potential for FastSAM technology in various real-world scenarios, such as industrial automation, medical imaging, and autonomous systems, where precise and efficient edge recognition is crucial. The codes will be released at \href{https://github.com/XFastDataLab/FastSmoothSAM}{$https://github.com/XFastDataLab/FastSmoothSAM$}}

\keywords{Image segmentation, Fast Segment Anything, B-Spline, Adaptive sampling}



\maketitle

\section{Introduction}\label{sec:introduction}

Image segmentation is a crucial task in computer vision, serving as a fundamental component of image processing. Accurate and precise segmentation is vital for a range of practical applications, including object detection\citep{object_detection1, object_detection3}, medical image analysis\cite{medical1}, and autonomous driving\cite{autonomous_driving1, autonomous_driving2}, among others. However, image segmentation remains challenging, particularly in complex scenes with irregular object shapes or variable lighting conditions. In this context, obtaining precise boundary information of targets is pivotal for advanced image understanding and applications. Consequently, tackling image segmentation challenges in environments with complex backgrounds, irregular contours, or subtle textures is crucial and necessitates the use of advanced algorithms and techniques. Continued enhancement of these algorithms, improving their adaptability to complex scenarios and challenging conditions, is critical to fulfilling the diverse requirements of real-world applications. This is particularly significant for the successful execution of tasks such as target recognition, scene understanding, and other related image processing activities.

The Segment Anything Model (SAM)\cite{SAM} is an unsupervised foundational model for image segmentation, particularly adept at zero-shot segmentation tasks. SAM excels as a tool for image segmentation, effectively managing various image types and achieving precise object segmentation. Through  deep learning of image feature distributions, SAM consistently performs object segmentation in complex environments, Its applications are widespread, including medical imaging\cite{MedSAM,RepViT-MedSAM,AdaptiveSAM}, natural scene analysis\cite{natural_scene}, and other areas. Distinct from traditional semantic segmentation methods, SAM incorporates engineering to optimize segmentation results with auxiliary information such as text, coordinates, and bounding boxes. This not only improves interaction flexibility but also represents a significant effort to address scalability challenges in image segmentation.

Due to the use of Vision Transformers (ViT)\cite{ViT1, ViT2, ViT3, ViT4} as the backbone in SAM, this approach leads to significant GPU memory consumption and slower inference speeds, which demand higher hardware capabilities. This becomes a notable limiting factor, especially in scenarios requiring rapid processing of large volumes of images or real-time applications. To address this challenge, a team from the Chinese Academy of Sciences introduced the FastSAM\cite{FastSAM}, an enhanced version of SAM. FastSAM maintains the segmentation accuracy of the original SAM algorithm while significantly boosting processing speeds—achieving speeds peak up to 50 times faster than SAM. With fewer parameters and reduced GPU memory usage, FastSAM is suitable for deployment in applications where precision demands are moderate, making it a viable alternative to SAM.

While FastSAM has made significant strides in improving processing speed, it still faces limitations when handling certain types of images, especially those with irregularly shaped objects or complex edges. A prevalent issue is FastSAM's tendency to produce jagged edges during segmentation. These jagged edges not only degrade the visual quality of the results but may also adversely affect subsequent image analysis and processing. This is particularly critical in scenarios that demand highly accurate edge information, such as precision medical imaging analysis and high-precision map creation.

Therefore, for these specific applications, further refinement of the segmentation algorithm is necessary to improve the quality and accuracy of the edges. Despite FastSAM's impressive processing speeds, its segmentation quality still requires further optimization and improvement in certain scenarios. This necessity has spurred the exploration of more efficient and precise image segmentation methods.

The B-Spline curve\cite{B-spline} is a curve and surface fitting technique that relies on the linear combination of control points and basis function to generate a smooth curve. This curve is determined by a sequence of control points, which offer high flexibility and adaptability. In the field of image processing, B-Spline curve can be utilized to accurately model the contours of objects within images\cite{B-spline1, B-spline2}, allowing for advanced shape manipulation and editing. These capabilities are particularly useful in tasks like object tracking and feature extraction, where precise and smooth representations are crucial. Additionally, the inherent smoothness of B-Spline curve makes them ideal for rendering anti-aliased graphics, smoothing out jagged lines to enhance visual quality. Furthermore, B-Spline curve fitting is very fast when applied to a limited set of sample points, especially using methods like the least squares technique. Consequently, this combination of speed and precision renders B-Spline fitting highly suitable for applications that require rapid image processing and real-time analysis.

\begin{figure}[htb]
  \center
  \includegraphics[scale=0.5]{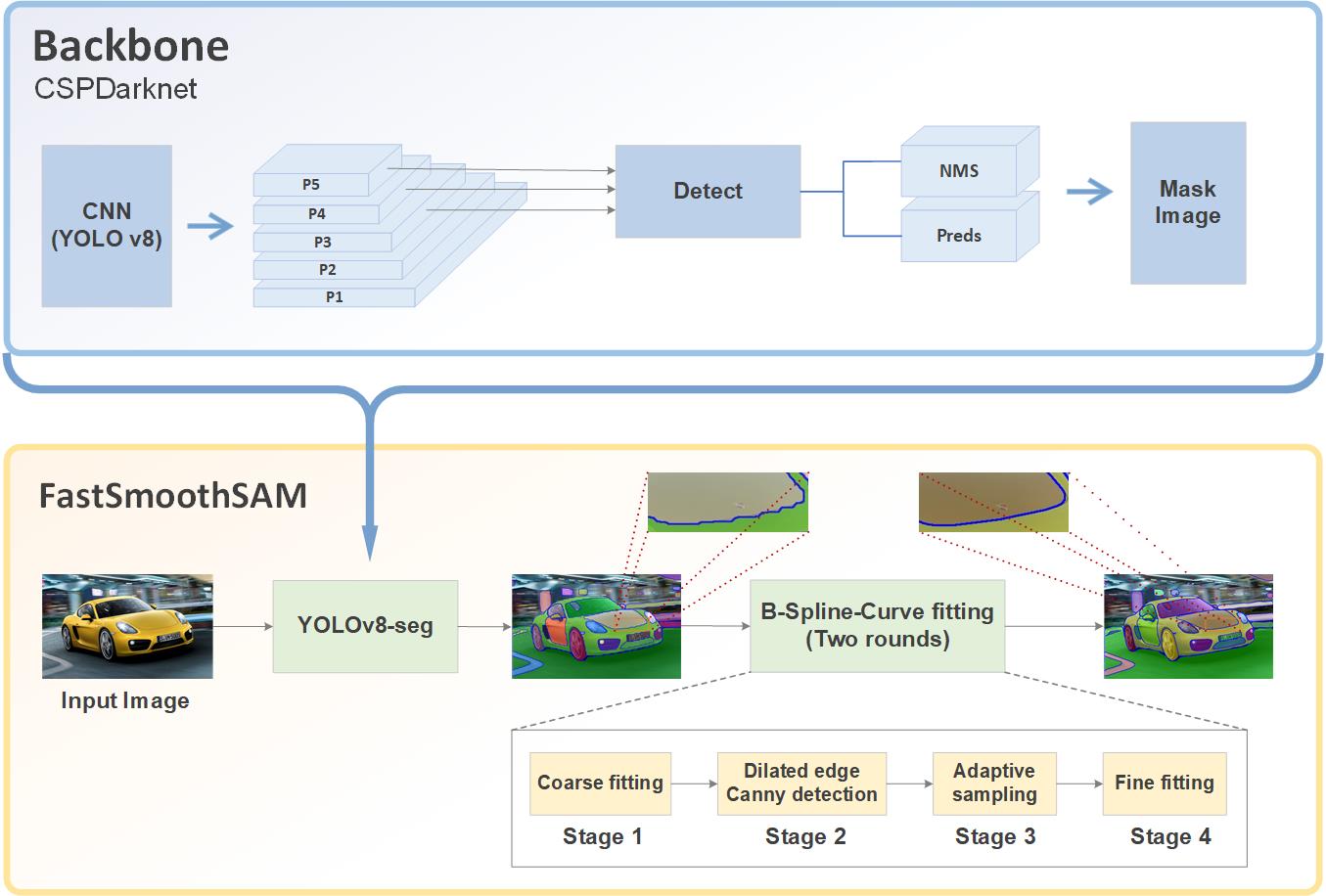}
  \caption{The framework of FastSmoothSAM. Our proposed FastSmoothSAM primarily utilizes YOLOv8-seg as the backbone for the image encoder (top), and depicted at the bottom is our operational framework. After employing two rounds B-Spline curve processing, this approach effectively addresses the issues of low accuracy and jagged edges inherent in the YOLOv8 network}
  \label{fig:framework}
\end{figure}

The primary aim of this study is to rapidly segment instances from images with high fidelity using YOLOv8-seg\footnote{\url{https://github.com/ultralytics/ultralytics}}. Fig. \ref{fig:framework} provides an overview of our methodology, termed FastSmoothSAM. Initially, the image is encoded, which may result in instances with jagged edges. To address this issue, we apply a four-stage refining process using two rounds of B-Spline curve fitting techniques to effectively smooth these edges. This approach significantly enhances the visual quality of the initial masks and better aligns them with the actual contours of the objects in the image.

Our main contribution can be summarized as follows:
\begin{itemize}
    \item We propose using \textbf{B-Spline Curve Fitting} to smooth the edges of objects identified by the CNN network.
    \item We introduce a \textbf{Two-Round-Smoothing Strategy}: the first round involves the estimation of the rough contours of the object edges initially, while the second round focuses on precise edge detection.
    \item  We develop a \textbf{Four-Stage-Edge-Curve-Fitting Method} to enhance edge segmentation: starting with smoothing rough contours through coarse fitting, then capturing Canny edges\cite{Canny} related to contour information, refining through adaptive sampling, and finally adjusting based on actual edge features for precise curve fitting.
\end{itemize}

\section{Related work}\label{sec:related works}

\subsection{Segment Anything Model}
The introduction of SAM\cite{SAM} marks the advent of a state-of-the-art image segmentation model in the field of computer vision. This innovation introduces a completely new image segmentation task, model, and dataset, representing a significant breakthrough for the field. The model's unique design philosophy is meticulously crafted to adapt to prompts, enabling flexible transfers to new image distributions and tasks in a zero-shot manner. In certain scenarios, SAM even outperforms traditional fully supervised learning methods.

Current research in image segmentation techniques primarily builds upon SAM, introducing improvements and adaptations for diverse tasks. Examples of these innovative adaptations include FastSAM\cite{FastSAM}, MobileSAM\cite{MobileSAM, MobileSAMv2}, EfficientSAM\cite{EfficientSAM}, SEEM\cite{SEEM}, HQ-SAM\cite{HQ-SAM}, and UnSAM\cite{UnSAM}. These variants not only enhance performance but also broaden the applicability of the algorithm, fueling significant advancements in image segmentation technology.

\subsection{Fast Segment Anything}
To meet the real-time demands of SAM, FastSAM has meticulously developed an innovative real-time solution. Distinct from other enhancements to the SAM algorithm, FastSAM adopts a unique approach. It introduces an instance segmentation branch based on a Convolutional Neural Network (CNN), built upon YOLOv8-seg, which combines the object detection capabilities of YOLO with the YOLACT\cite{Yolact} method. This specific design is optimized to address panoramic segmentation tasks, ensuring efficient and effective performance.

FastSAM divides the entire segmentation task into two sequential stages: all-instance segmentation and prompt-guided selection. Firstly, in the all-instance segmentation stage, a CNN-based detector like YOLOv8-seg efficiently segments all objects or regions in the image, generating the corresponding segmentation masks. Following this, in the second stage, the model outputs the regions of interest corresponding to the prompts. By harnessing the computational efficiency of CNN, FastSAM successfully demonstrates the feasibility of achieving real-time segmentation without compromising performance quality. This innovative approach provides robust support for meeting the demands of real-time segmentation tasks, and the unique design and successful implementation of real-time performance by FastSAM open new avenues for addressing the need for immediate and effective image segmentation in practical applications.

However, a major flaw of FastSAM is that it often produces unsmooth object edges. This roughness can compromise the fidelity of the segmentation, negatively impacting the performance of downstream tasks such as object recognition and tracking.

\subsection{B-Spline Curve}
B-Spline curve represents a powerful technique for curve and surface fitting, formed by the linear combination of control points and basis functions. These curves are highly adaptable, owing to the ability to modify the positions and weights of the control points, thus offering fine control over the curve's shape. B-Spline curve utilizes optimization methods that aim to minimize approximation errors, ensuring that the curve tightly conforms to the given data points. This accuracy and smoothness make B-Spline a preferred choice in many fields that require precise data representation.

B-Spline are extensively utilized in image processing for various tasks, including image   deformation\cite{deformation1, deformation2}, restoration\cite{restoration2} and geometric design\cite{THB-splines}. By employing B-Spline curve within images, their inherent flexibility facilitates smooth transformations, region selection, boundary definition, missing part reconstruction, and shape adjustments. These capabilities make B-Spline effective tools for intricate image manipulation tasks.

\section{The basic of B-Spline curve}
\subsection{The definition of B-Spline curve}
Given $n+1$ control points $P=[P_0, P_1, \ldots, P_n]^T$ ($P_i \in R^{2\times 1}$) and knot vector $U=\{u_0,u_1,\ldots, u_m\}$, the expression of a B-Spline curve is:

\begin{equation}
C(t)=\sum_{i=0}^{n}N_{i,k}(t)P_i,
\label{equ:bspline}
\end{equation}
where 

\vspace{4pt}
$t\in [u_0,u_m]$;

\vspace{4pt}
$k$ is the degree of the curve ($k$ = 1, 2, 3, \ldots);

\vspace{4pt}
$m=n+k+1;$

\vspace{4pt}
$C(t)$ is the coordinate vectors of points on the curve;

\vspace{4pt}
$N_{i,k}(t)$ is the B-spline basis function, it can be constructed using the $Cox–de Boor$ recursion formula:
\begin{equation}
\label{equ:basis_functions_0}
{
N_{i,0}(t)=\left\{ 
    \begin{aligned}
    1 & \quad if \quad u_i \leq t \leq u_{i+1}\\
    0 & \quad\quad otherwise \\
    \end{aligned}
    \right
    .
} .
\end{equation}

% \begin{equation}
\begin{align}
\label{equ:basis_functions_k}
N_{i,k}(t) = \frac {t-u_i}{u_{i+k}-u_i}N_{i,k-1}(t) + \frac {u_{i+k+1}-t}{u_{i+k+1}-u_{i+1}}N_{i+1,k-1}(t).
\end{align}
% \end{equation}
% N_{i,k}(t) &= \frac {t-u_i}{u_{i+k}-u_i}N_{i,k-1}(t) 
% \notag
% \\ &+ \frac {u_{i+k+1}-t}{u_{i+k+1}-u_{i+1}}N_{i+1,k-1}(t)

The derivative of $C(t)$ can be computed as follows:
\begin{equation}
\label{eq:der}
    C'(t)=\frac{d}{dt} (C(t))=\sum_{i=0}^{n-1} N_{i+1,k-1}(t) Q_i,
\end{equation}
where $Q_i=\frac{k}{u_{i+k+1}-u_{i+1}}P_{i+1}-P_i$.

\vspace{4pt}
Similarly, the second-order derivative of $C(t)$ is as follows:
\begin{equation}
\label{eq:2ndder}
    C''(t)=\frac{d}{dt} (C'(t))=\sum_{i=0}^{n-1} N_{i+1,k-2}(t) D_i,
\end{equation}
where $D_i=\frac{k}{u_{i+k}-u_{i+1}}Q_{i+1}-Q_i$.

\vspace{4pt}
The curvature of $C(t)$ is:
\begin{equation}
\label{eq:curv}
    K=\frac{\|C'(t)\times C''(t)\|}{\|C'(t)\|^3}.
\end{equation}

\subsection{Approximate B-Spline fitting}
Given a set of $a+1$ data points, $D_0, D_1, ..., D_a$, and a degree $k$ with unknown control points, suppose there 
exists a B-spline curve of degree $k$ defined by $b+1$ control points $P_0,P_1, ..., P_b$, such that the curve approximately passes all data points in the given order with parameters $t_0, t_1, ..., t_a$ where $U_0=t_0 <t_1 < ...<t_{a-1} <t_a=U_m$, i.e, : 

\begin{equation}
D=NP
\label{equ:D=NP},
\end{equation}
where 

%C(t)=\sum_{i=0}^{l}N_{i,p}(t)P_i,
\vspace{4pt}
$P=[P_0, P_1, ..., P_b]^T$;

\vspace{4pt}
$D_i=\sum_{i=0}^{b}N_{i,k}(t_i)P_i$;

\vspace{4pt}
$D=[D_0, D_1, ..., D_a]^T$;

\vspace{4pt}
$
N=
  \left[
    \begin{array}{cccc}
     N_{0,k}(t_0) & N_{1,k}(t_0) & \cdots & N_{b,k}(t_0) \\
     N_{0,k}(t_1) & N_{1,k}(t_1) & \cdots & N_{b,k}(t_1) \\
     \vdots & \vdots & \ddots & \vdots \\
     N_{0,k}(t_a) & N_{1,k}(t_a) & \cdots & N_{b,k}(t_a) \\
    \end{array}
  \right]
$
\vspace{4pt}

Then, the control point set $P$ can be approximately deduced by the least squares method:

\begin{equation}
\label{equ:ls}
P={(N^{T}N)}^{-1}N^{T}D.
\end{equation}

The algorithm for the B-Spline curve approximately fitting used in this paper is as Alg. \ref{alg:bspline_fitting} with the complexity $O(a^3)$ where $a$ is the number of data points number in $D$. However, in our applications $a$ is small, it can run in about constant expected time $O(C)$.
\begin{algorithm}
  \setstretch{1.4}
  \footnotesize
  \caption{B-Spline curve fitting} 
  \label{alg:bspline_fitting} 
  \begin{algorithmic}
    \Statex \textbf{Input:} 
    Data points $D=\{D_0, D_1,..., D_a\}$;
    
    Degree $k$;
    
    Knot vector $U={u_0, u_1,..., u_m}$.
    \Statex \textbf{Output:} Control points $P$
    \Statex \textbf{Step 1:} Generate uniform parameter: $t_i=u_0+(i-1)\frac{u_m-u_0}{a}$
    \Statex \textbf{Step 2:} Create matrix $N$
    \Statex \textbf{Step 3:} $P={(N^{T}N)}^{-1}N^{T}D$
    \Statex \textbf{Return} $P$     
\end{algorithmic} 
\end{algorithm}

\section{The Proposed Method}
\label{sec:Method}
\subsection{Motivation}
\begin{figure}[htbp]
  \centering
  % \subfigure[Original Image]
  {\includegraphics[height=2.2in]{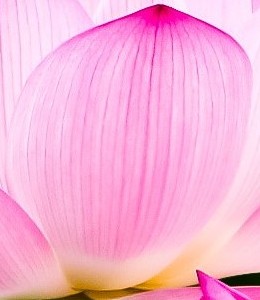}}
  \hspace{4mm}
  % \subfigure[Segmentation by YOLOv8]
  {\includegraphics[height=2.2in]{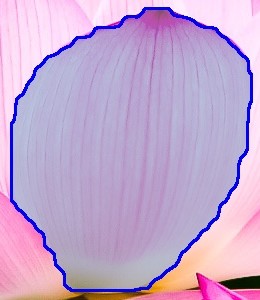}} \\
  \scriptsize ~~~~~~~~~(a) Original image ~~~~~~~~~~~~~~~(b) Segmentation by YOLOv8
  \caption{A poor outcome in edge identification by YOLOv8 occurs when parameters are not appropriately chosen}
  \label{fig:yolo_jagged_edge}
\end{figure}

Current methodologies, like FastSAM, accelerate the process of image segmentation by optimizing the YOLOv8 framework instead of ViT\cite{ViT1}. Nevertheless, these methods often struggle to deliver the desired accuracy in complex environments or when handling images with irregular edges. The results frequently suffer from jagged edges, a common issue with YOLOv8 when it fails to capture precise boundary details due to its inherent limitations in feature extraction and the effects of downsampling.  Fig. \ref{fig:yolo_jagged_edge}(a) illustrates a petal, and (b) displays a segmentation of the petal with a jagged edge, as obtained by YOLOv8.

In scenarios where precise edge definition is essential, there is a clear need for further enhancements in edge detection and processing techniques within this framework. These issues can be addressed by introducing edge-aware loss functions, employing data augmentation techniques to enhance edge features, and adjusting the network architecture. However, implementing these solutions can be costly and complex. Therefore, utilizing post-processing techniques presents a more feasible option. As mentioned earlier, B-Spline curve fitting offers significant advantages and is particularly well-suited for smoothing jagged edges in segmented images. This approach not only improves the aesthetic and accuracy of the segmentation results but also does so cost-effectively, making it an attractive solution for refining edges in image segmentation tasks.

Therefore, our goal is to adopt a post-processing approach that allows for rapid and cost-effective image smoothing and segmentation. Specifically, we use FastSAM\cite{FastSAM}, a lightweight initial image segmentation model that leverages YOLOv8 on a relatively small dataset, to identify the rough contours of an object. Following this, we employ B-Spline curve fitting to refine and smooth the edges of the segmented images. This two rounds process ensures high-quality segmentation results, even with limited training data. The use of B-Spline fitting as a post-processing step not only enhances the visual quality of the edges but also significantly reduces the overall computational burden and cost, making this approach highly practical for various applications.

\subsection{Edge smoothing based on B-Spline}
\label{subsec:main_stage}
Using B-splines for object edge fitting involves several detailed processes to ensure optimal results. These processes include edge extraction, sorting, dynamic sampling, and multi-round fitting. In this paper, we propose the \textbf{Four-Stage-Edge-Curve-Fitting Method}, which ensures accurate contour representation by carefully managing the order and density of data points, followed by iterative optimization. This comprehensive approach is crucial for achieving precise and reliable edge-fitting results. The main processes are plotted in Fig. \ref{fig:four_stage_edge_curve_fitting}, and the details are presented below.

\begin{figure}[htb]
  \centering
  \includegraphics[width=5.8in]{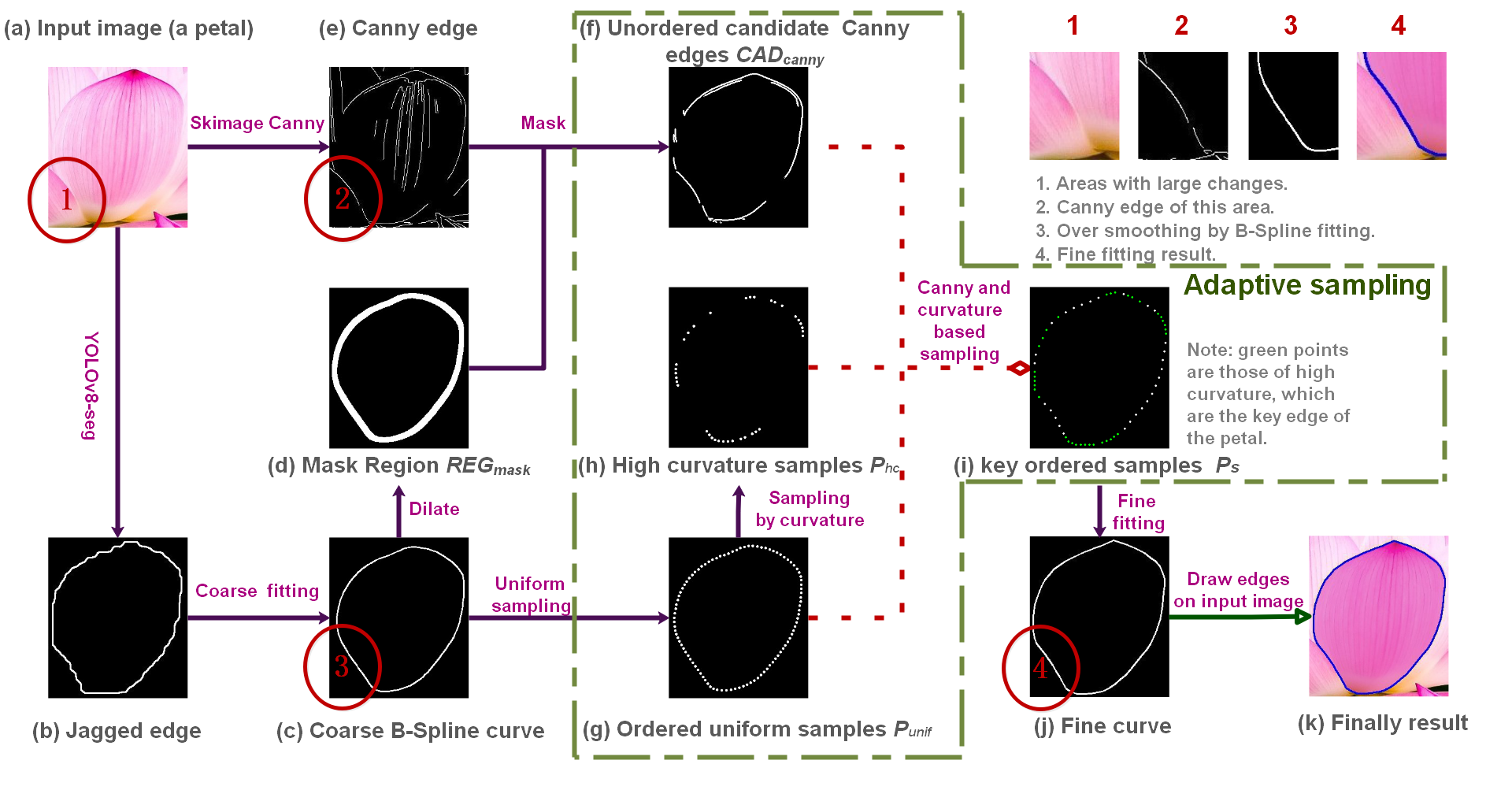}
  \caption{The main processes of edge fitting based on two rounds of B-Spline fitting, where (c) is the initial coarse fitting, and (j) is the fine fitting after adaptive sampling which comprehensively takes Canny edge and curvature into consideration. The red circle 1, 2, 3 and 4 highlight an optimized local area. Compared to circle 2, the fitting curve within circle 4 is more consistent with the edge of the petal within circle 1}  \label{fig:four_stage_edge_curve_fitting}
\end{figure}

% \textbf{Stage 1 Coarse fitting:}
\subsubsection{Stage 1: Coarse fitting}
% \label{subsec:Stage 1}

Given the contour of an object identified by YOLOv8 or other algorithms, fitting B-spline curve to the contour involves several key steps to ensure accurate and precise contour representation. Here are the main steps:

1) Extract Ordered Edge Points: Extract edge data points from the contour which saves the edges point in order.

2) B-Spline Curve fitting: Perform the fitting using the least squares method or other suitable optimization techniques to achieve a precise representation of the contour.

B-Spline fitting is performed on edge data points using Alg. \ref{alg:bspline_fitting} with a degree of 3.

However, during our experiments, we observed that direct curve fitting often yields suboptimal results. As indicated by the red circle 1 and 3 of Fig. \ref{fig:four_stage_edge_curve_fitting}, the fitted edges do not align well with the actual contours of the petal, especially in areas of high curvature, the curve tends to be overly smooth.

% \textbf{Stage 2 Dilated edge Canny detection:}
\subsubsection{Stage 2: Dilated edge Canny detection}
% \label{subsec:Stage 2}

As we can see, the results of coarse fitting may not align consistently with the actual edges of the object. Therefore, it is essential to consider the original contours of the object to restore key details.

In this paper, the Skimage Canny algorithm is utilized to optimize gradient calculation and edge tracking. Additionally, adjusting the parameter $\sigma$ for Gaussian smoothing filtering enhances edge extraction (see Eq. (\ref{equ:gassian})), which effectively mitigates poor edge detection performance caused by subtle color differences.

\begin{equation}
\label{equ:gassian}
    G(x,y)=\frac{1}{2\pi\sigma^2}EXP(-\frac{x^2+y^2}{2\sigma^2}).
\end{equation}

There are two key operations in this stage:

1) Dilate the curve from the coarse fitting to obtain a mask edge region, denoted as $REG_{mask}$.

2) Apply the Skimage Canny algorithm to identify Canny edge points within $REG_{mask}$, referred to as Candidate Canny Edges $CAD_{canny}$. 

For example, as shown in Fig. \ref{fig:four_stage_edge_curve_fitting}(c), (d), (e) and (f):

(c) displays the coarse edge fitting result of the petal. 

(d) shows the dilated region from (c), which is used as a mask. 

(e) depicts the Canny edges derived from the original image. 

(f) illustrates the unordered Canny edge points located within the mask $REG_{mask}$.

% \textbf{Stage 3 Adaptive sampling:}
\subsubsection{Stage 3: Adaptive sampling}
% \label{subsec:Stage 3}

In Stage 2, we obtain the Canny edge points within the dilated region, which are presumed to closely represent the true edges of the object. However, directly fitting B-Spline curve by these points presents two problems:

1) B-Spline fitting demands ordered data points, whereas the edge data points obtained are unordered.

2) Employing the least squares method for B-Spline fitting often leads to over-smoothing, particularly in areas of high curvature. This smoothing effect can cause the curve to deviate significantly from the actual object edges. Consequently, areas with high curvature require a denser concentration of data points compared to areas with lower curvature to maintain edge accuracy.

To address this issue, we implemented an adaptive sampling strategy on object edges. This method strategically allocates more sampling points in areas with higher curvature—where the object boundaries have greater complexity and detail—while reducing the number of samples in smoother regions. The process involves three sub-steps, outlined as follows:

\textbf{Curvature-based sampling :} We focus on extracting edge points specifically within regions of high curvature, which are crucial for roughly sketching the contour of the object. This approach is illustrated in Fig. \ref{fig:four_stage_edge_curve_fitting}(h). The details of this technique are outlined in Alg. \ref{alg:curvature_based_sampling}, with a computational complexity of approximately  $O(n)$, where $n$ represents the number of sampling knots.

\begin{algorithm}[htbp]
  \setstretch{1.4}
  \footnotesize
  \caption{Curvature-based sampling} 
  \label{alg:curvature_based_sampling} 
  \begin{algorithmic}[1]
   \Require 
    Controls points $P$ of a B-Spline curve $S$; 

    % Degree $k$ of B-Spline curve $S$
    
    Knots vector $U$ of $S$;

    The number of sampling points $n$;
    
    Curvature threshold $\theta$.
    \Ensure point set $P_{hc}$ of high curvature
    \State $T$ = uniformly sample $n$ number of knots t from $U$ 
    \For{each $t \in T$}
        \State compute $C'(t)$ and $C''(t)$ by Eq. (\ref{eq:der}) and (\ref{eq:2ndder}) 
        \State compute curvature $K$ by Eq. (\ref{eq:curv})
        \If {$K > \theta$ }
        \State add $C(t)$ point $p_i$ into $P_{hc}$
        \EndIf
    \EndFor
    \State \textbf{Return} $P_{hc}$
\end{algorithmic} 
\end{algorithm}

\textbf{Canny-based sampling :} The Canny edge points $CAD_{canny}$ obtained from Stage 2 within $REG_{mask}$ still present several challenges: 1) some contours are often missing, compromising the completeness of the edge mapping, and 2) noise is included, which can significantly affect the accuracy of the subsequent B-Spline fitting.  

To address these issues, we propose a simple strategy to refine sampling for further B-Spline fitting. This involves taking into account the coarse fitting results, the unordered Canny edge points $CAD_{canny}$ and key points of high curvature. Specifically, we resample according to the order of the initial uniform sample $P_{unif}$, and for each sample point $p_i \in P_{unif}$, there are three cases as follows and more details are shown in Alg. \ref{alg_Canny_based_sampling}:

\begin{algorithm}
  \setstretch{1.4}
  \footnotesize
  \caption{Canny-based sampling} 
  \label{alg_Canny_based_sampling} 
   \begin{algorithmic}[1]
    \Require $P_{unif}$: Uniform coarse fitting sample point set ; 
    
    $P_{hc}$: High curvature point set ; 
    
     $CAD_{canny}$: Canny edge points within $REG_{mask}$; 
    
    $r$: Searching radius of KD tree.
    \Ensure $P_{s}$: Key ordered sample 
    \State $P_s = \emptyset$
    \State $label = \{0,0,...,0 \}$
    \Comment{denote all samples are not handled} 
    \For{each $p_i \in P_{unif}$}
        \State //Neighbor query within $r$-neighborhood by K-D tree.
        \State NNCannySet= $NN(p_i,CAD_{canny},r)$ 
        \State //Missing Canny edge
        \If {$NNCannySet$ is empty}  
        \Comment{Case 1}
        \State Add $p_i$ into $P_s$
        
        \ElsIf {$p_i \in P_{hc}$}
        \Comment{Case 2}
        \State $p_{cent}$ = the center of $NNCannySet$
        \State Add  $p_{cent}$ into $P_{s}$
        
        \ElsIf{$label[i]==0$} 
        \Comment{Case 3 $p_i$ is not handled}
        \State NNCoareSet= $NN(p_i,P_{unif},r)$
        \State $idsOfNN$= the indexes of NNCoareSet in $P_{unif}$ 
        \State //{label all elements in NNCoareSet are handled.}
        \State $label[idsOfNN]=1$ 
        \State $p_{cent}$ = center of $NNCannySet \bigcup NNCoareSet$
        \State Add $p_{cent}$ into $P_{s}$
        \EndIf
        %\State $label[i]=1$ 
    \EndFor
    \State \textbf{Return} $P_s$
\end{algorithmic} 
\end{algorithm}

\textbf{Case 1:} In areas where $CAD_{canny}$ edges are missing, we directly use the $p_i$.

\textbf{Case 2:} In areas of high curvature, only $CAD_{canny}$ is considered. Specifically, the center of the local $CAD_{canny}$ within an $r$-neighborhood of $p_i$, denoted as $NN(p_i, CAD_{canny},r)$, is selected. This approach ensures that samples of high curvature are prioritized.

\textbf{Case 3:} In other areas, we compute the center by considering both $NN(p_i, CAD_{canny},r)$ and $NN(p_i, P_{unif},r)$. This strategy reduces the number of samples in areas of low curvature, focusing on essential points for curve fitting.

The rationale behind this strategy is as follows: as mentioned above, B-Spline fitting on uniform samples may result in over-smoothing in areas of high curvature. To counteract this, one solution is to decrease the density of samples in areas of low curvature while maintaining the density in areas of high curvature, This approach effectively increases the relative weight of high curvature areas in the fitting process.

The time complexity of this strategy is approximately $O(n \times \log(n))$, where $n$ is the number of uniform sampling points. The function $NN(p,P,r)$ identifies neighbors of $p$ within its $r$-neighborhood in $P$ by K-D tree\cite{KD-tree, chen}. Fig. \ref{fig:four_stage_edge_curve_fitting}(i) presents an example of the resampling results: the green points represent areas of high curvature with high density, while the white points are sparse, indicating the relative flatness of these curve regions.

By adaptively adjusting the sampling results, we achieve a more accurate representation of the object’s true shape. This precision is particularly important in applications where edge integrity is critical. Adaptive sampling not only enhances the visual quality of the segmented images but also bolsters the robustness of subsequent analyses. This makes the overall system more efficient and effective.

% \textbf{Stage 4 Fine fitting:} 
\subsubsection{Stage 4: Fine fitting} 
% \label{subsec:Stage 4}

B-Spline fitting is performed again on the key ordered samples $P_s$ using Alg. \ref{alg:bspline_fitting} with a degree of 2. 

\subsection{Overall algorithm}
As mentioned above, Alg. \ref{alg:overall} presents the overall algorithm of the proposed method.

\begin{algorithm}
  \setstretch{1.4}
  \footnotesize
  \caption{FastSmoothSAM} 
  \label{alg:overall} 
   \begin{algorithmic}[1]
    \Require Image $I$; 
    \Ensure Masked image $I_{mask}$
    \State $MaskSets$ = Segment $I$ by YOLOv8-seg
    \For{each mask $M_i \in MaskSets$}
    \State /*\textbf{Stage 1: Coarse fitting}--------------------------------------------*/
    \State $EdgPoints$ = Extract edges from the contour of $M_i$
    \State $C_1$ = Fit a curve by Alg. \ref{alg:bspline_fitting} for $EdgPoints$
    \State /*\textbf{Stage 2: Dilated edge Canny detection}--------------------*/
    \State $REGmask$ = Dilate the region of $C_1$.
    \State $CannyEdges$ = Get Canny edge by $Skimage\ Canny$ 
    \State $CAD_{canny}$ = Mask $REG_{mask}$ on $CannyEdges$.
    \State /*\textbf{Stage 3: Adaptive sampling}------------------------------------*/
    \State $KeyEdges$ = Sampling by Alg. \ref{alg:curvature_based_sampling} from $C_1$
    \State  $P_s$ = Sampling by Alg. \ref{alg_Canny_based_sampling} from $C_1$, $CAD_{canny}$ and $KeyEdges$.
    \State /*\textbf{Stage 4: Fine fitting}------------------------------------------------*/
    \State $C_2$ = Fit a new curve by Alg. \ref{alg:bspline_fitting} for $P_s$
    \State $I_{mask}(i)$ = Obtain new mask by $C_2$
        %\State $label[i]=1$ 
    \EndFor
    \State \textbf{Return} $I_{mask}$
\end{algorithmic} 
\end{algorithm}

\textbf{Complexity analysis:} 
Let $N$ be the pixel number of image $I$, $N_i$ be the pixel number in mask $M_i$, $n_i$ be the number of sample points sampling from $i$-th contour, and $MaskNum$ be the number of masks.  For each loop from line 2 to line 16, we have:

In Stage 1: the operation of line 4 is to call $FindContours$ function in $OpenCV$ which has linear complexity $O(N_i)$, and Alg. \ref{alg:bspline_fitting} runs in about $O(n_i^3)$. 

In Stage 2: the main cost line in line 8 which calls $Skimage\ Canny$ with complexity about $O(N_i)$.

In Stage 3: the complexities of Alg. \ref{alg:curvature_based_sampling} (line 11) and Alg. \ref{alg_Canny_based_sampling} (line 12) are about $O(n_i)$ and $O(n_i \times \log(n_i))$, respectively.

In Stage 4: line 14 and line 15 have complexities about $O(n_i^3)$ and  $O(N_i)$, respectively.

Overall, the comprehensive complexity is about $O(MaskNum\times \Sigma(N_i+n_i^3))$, and usually due to that $N>>n_i$, FastSmoothSAM runs in about $O(N)$ expected time after segmenting by YOLO. 

\section{Experiments}
\label{sec:experiments}
In this section, we evaluate the performance of FastSmoothSAM across various application scenarios, analyzing both its advantages and limitations. We showcase visualizations of FastSAM’s segmentation using $everything$ modes, and compare these results with those of SAM and the ground truths.

\subsection{Experimental Setup}
We implemented and evaluated FastSmoothSAM using the Python programming language and the PyTorch library on a 2.10GHz 12th Gen Intel(R) Core(TM) i7-12700 machine running Windows, equipped with 32GB of memory+ Nvidia GTX 3060Ti 8GB and 3090 24G graphics cards.

\subsection{Visualized experiments}
\begin{figure}[htbp]
  \centering
  \subfigure[Inport image]{\includegraphics[width=1.38in, height=1.38in]{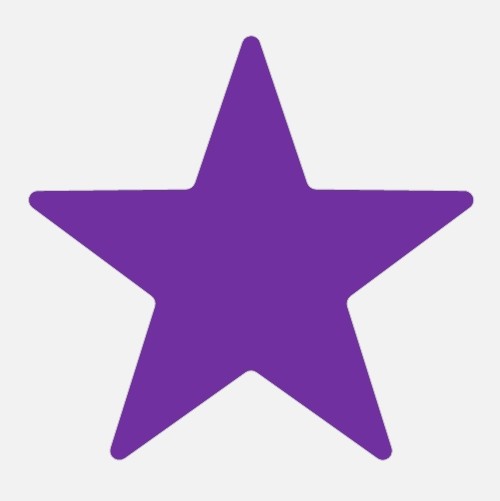}}
  \subfigure[YOLOv8]{\includegraphics[width=1.38in, height=1.38in]{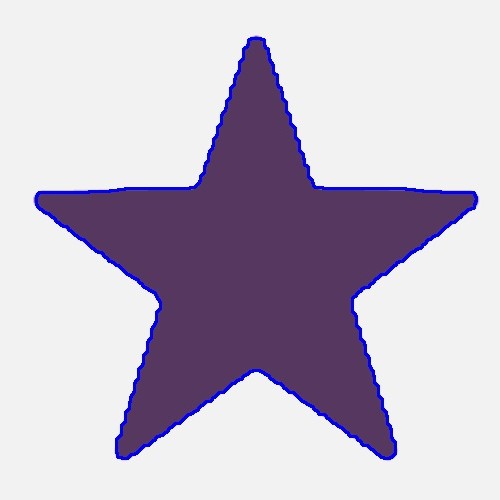}}
  \subfigure[Coarse fitting]{\includegraphics[width=1.38in, height=1.38in]{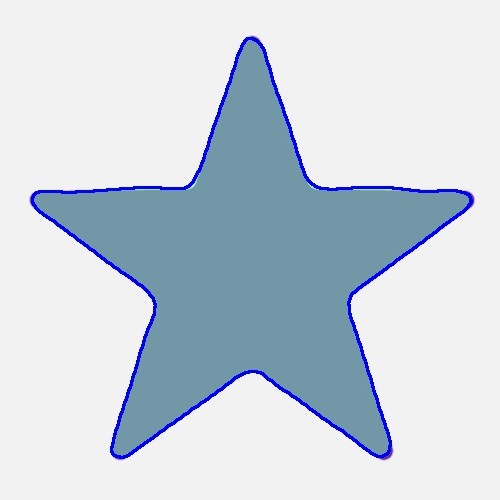}}
  \subfigure[High curvature]{\includegraphics[width=1.38in, height=1.38in]{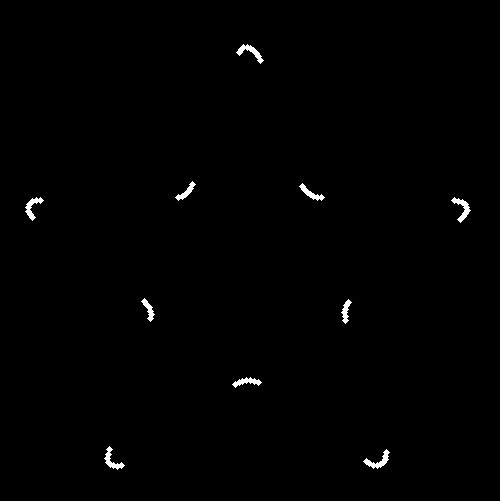}}
  \subfigure[Canny edge]{\includegraphics[width=1.38in, height=1.38in]{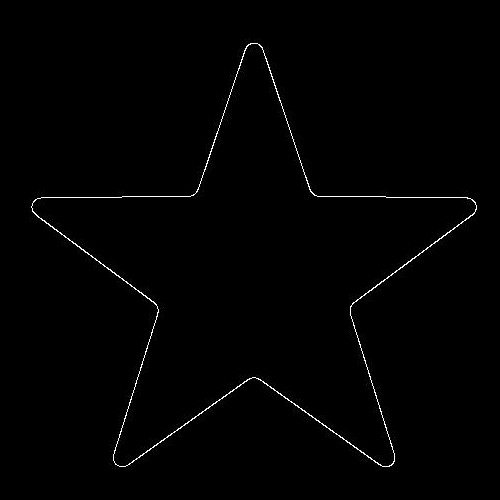}}
  \subfigure[Key edge]{\includegraphics[width=1.38in, height=1.38in]{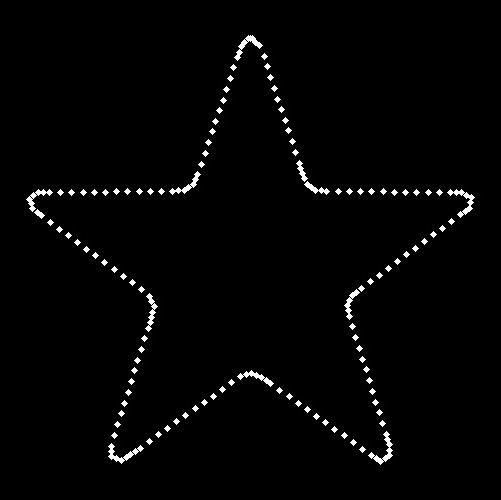}}
  \subfigure[Fine fitting]{\includegraphics[width=1.38in, height=1.38in]{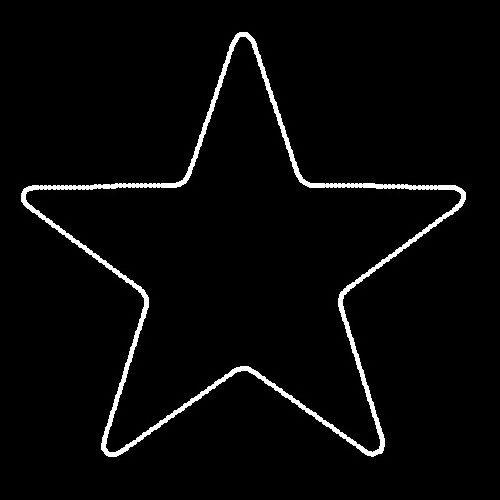}}
  \subfigure[Visualization]{\includegraphics[width=1.38in, height=1.38in]{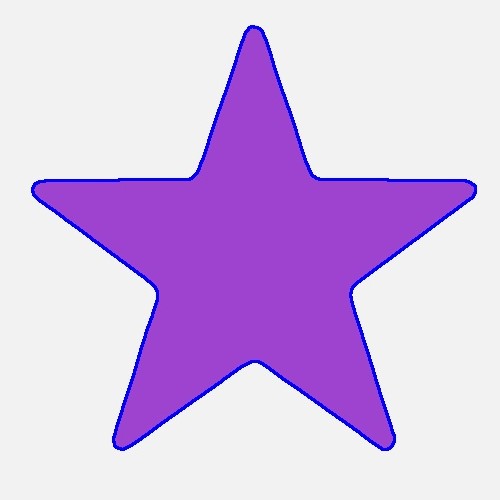}}
  \centering
  \caption{An example of displaying the visualization results of the FastSmoothSAM algorithm process on a pentagram}
  \label{fig:process_visualization}
\end{figure}

We conducted an experiment on a pentagon. Through this experiment, we can intuitively and systematically evaluate the segmentation quality and performance of the proposed method. The evolution process of the effect of the method in this paper is shown in Fig. \ref{fig:process_visualization}, where 

(a) The input image of a pentagram.

(b) Results are directly obtained from the YOLOv8-seg image encoder, where noticeable jagged edges are observed.

(c) The outcome after the first B-Spline curve fitting through the coarse fitting.

(d) A curvature-based sampling of the coarse fitting results to preserve key information for subsequent edge points.

(e) Edge point information obtained by the Skimage Canny algorithm, aims to closely match the original edges.

(f) Results are obtained by combining the coarse fitting result, the precise edges $CAD_{canny}$, and key points of high curvature.

(g) The final outcome after the second B-Spline curve fitting through the fine fitting, which preserves essential information and refines the edge fitting curve.

(h) Visualization of the final result.

It is evident that, compared to the results obtained directly from YOLOv8, our method produces more accurately edges fit the input image.

\subsection{Comparisons experiments}

\subsubsection{Visualized comparisons}
\begin{figure}[htbp]
  \centering
  % \rotatebox{90}{\scriptsize{~~~~~~~Ours~~~~~~~~~~~~~~~~FastSAM~~~~~~~~~~~~~~~~~Input}}
  \includegraphics[width=5.6in]{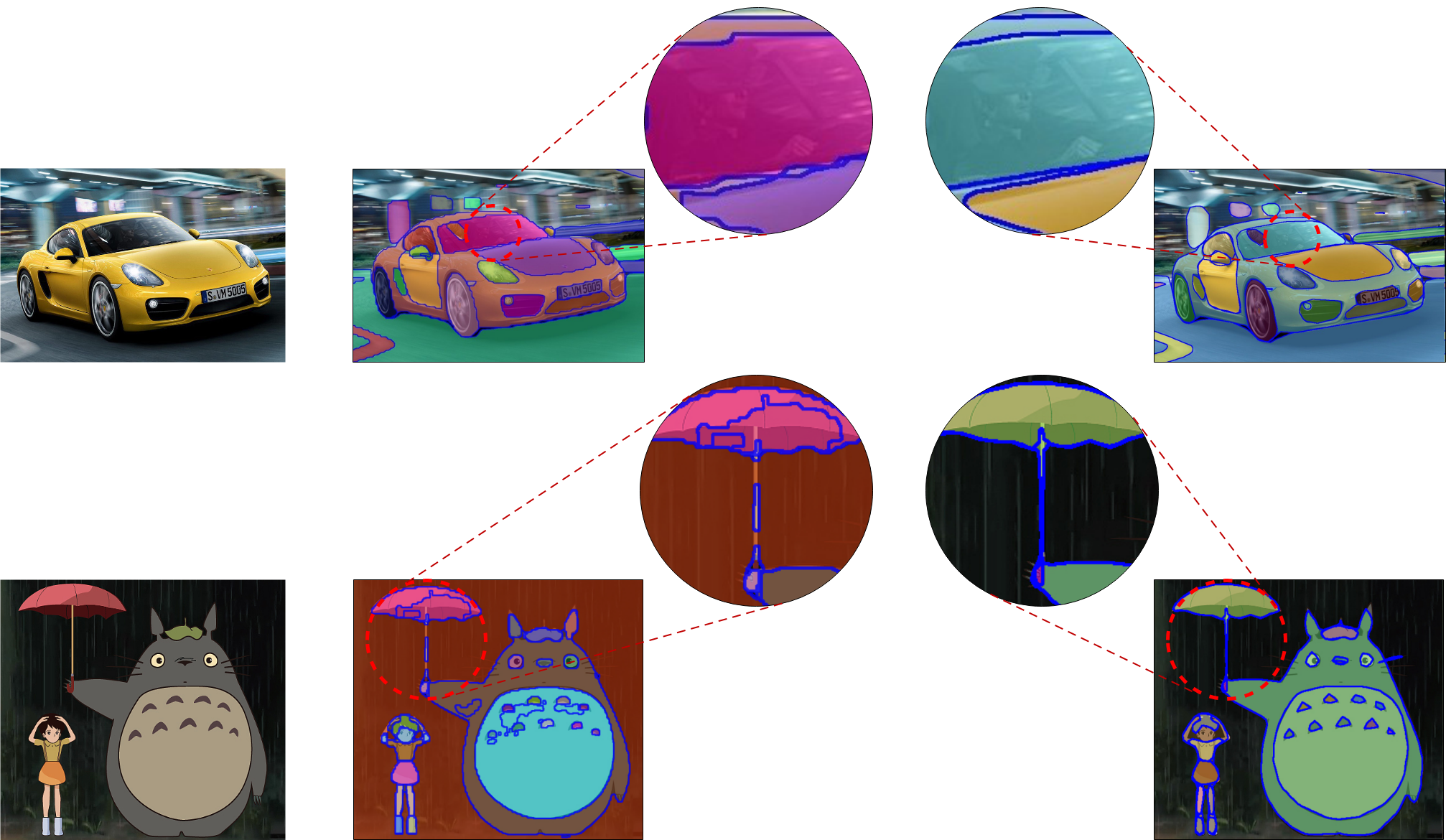}
  % \\(a) Input image~~~~~~~~~~~~~~~~~~~~~~~(b) FastSAM~~~~~~~~~~~~~~~~~~~~~~(c) Ours
  \\\leftline{\footnotesize \hspace*{0.3in}Input image \hspace{1.1in} FastSAM \hspace{1.7in} FastSmoothSAM}
  \caption{Comparison of experimental results between FastSmoothSAM and FastSAM. The left image represents the original image, the middle image depicts the experimental results from FastSAM, which exhibit jagged edge features, and the image on the right shows the experimental results from FastSmoothSAM, which demonstrate significantly smoother contour edges}
  \label{fig:vis_comparison}
\end{figure}

The experiment conducted on two images, as shown in Fig. \ref{fig:vis_comparison}, reveals contrasting results. FastSAM exhibits jagged edge features, whereas our proposed method accurately captures smooth contour edges. This observation underscores the efficacy of our method in achieving superior image segmentation.

\begin{figure}[htbp]
  \centering
  \subfigure{\includegraphics[width=1.38in, height=1in]{Fig6-1.jpg}}
  \subfigure{\includegraphics[width=1.38in, height=1in]{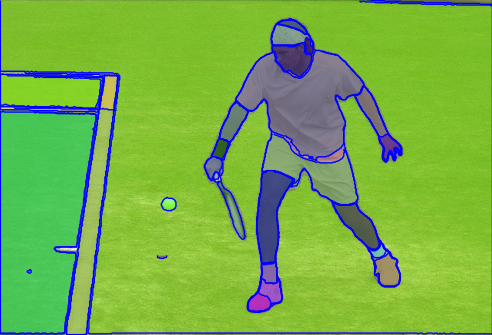}}
  \subfigure{\includegraphics[width=1.38in, height=1in]{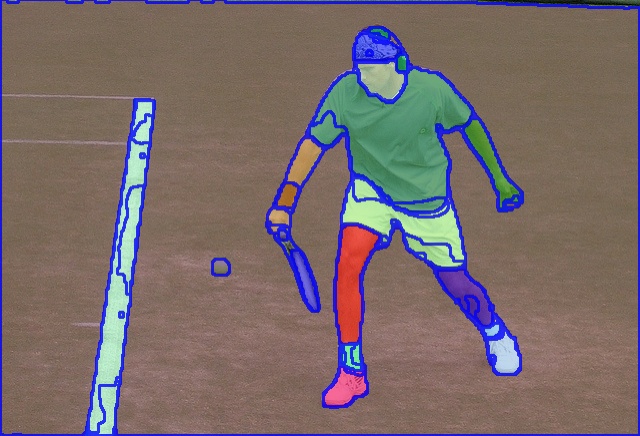}}
  \subfigure{\includegraphics[width=1.38in, height=1in]{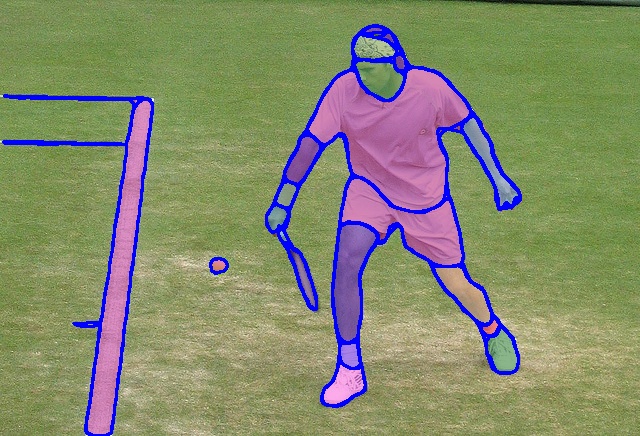}}
  \subfigure{\includegraphics[width=1.38in, height=1in]{Fig6-5.jpg}}
  \subfigure{\includegraphics[width=1.38in, height=1in]{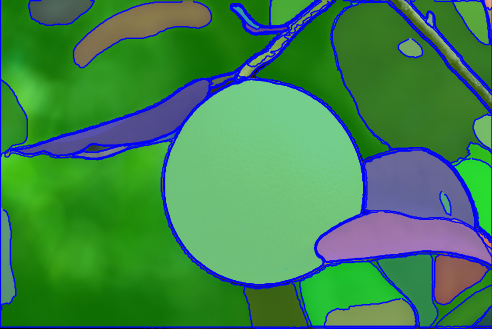}}
  \subfigure{\includegraphics[width=1.38in, height=1in]{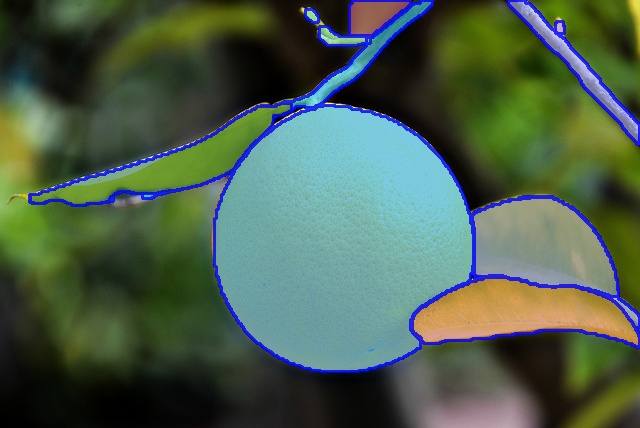}}
  \subfigure{\includegraphics[width=1.38in, height=1in]{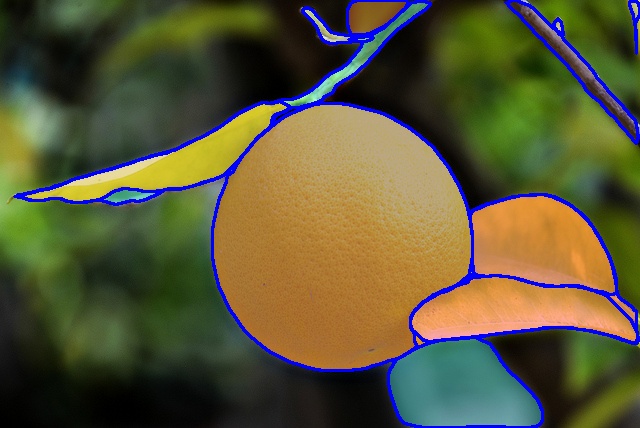}}
  \subfigure{\includegraphics[width=1.38in, height=1in]{Fig6-9.jpg}}
  \subfigure{\includegraphics[width=1.38in, height=1in]{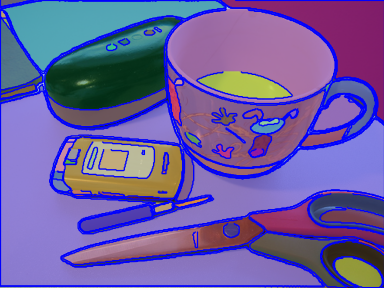}}
  \subfigure{\includegraphics[width=1.38in, height=1in]{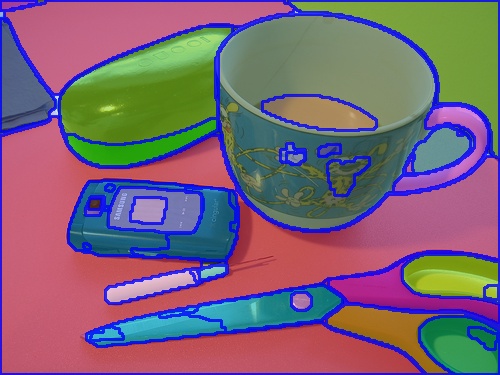}}
  \subfigure{\includegraphics[width=1.38in, height=1in]{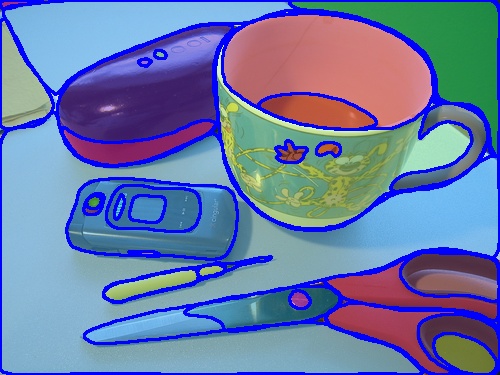}}
  \\\leftline{\scriptsize \hspace*{0.5in}Image \hspace{1.1in} SAM \hspace{0.9in} FastSAM \hspace{0.6in} FastSmoothSAM}
  \centering
  \caption{Comparison of segment everything results.}
  \label{fig:exp_more}
\end{figure}

As shown in Fig. \ref{fig:exp_more}, comprehensive comparative experiments were conducted on the COCO dataset\footnote{\url{https://github.com/cocodataset/cocoapi}}\cite{coco}. The experimental results demonstrate that, compared to other methods, the FastSmoothSAM model maintains stable performance across various complex scenarios and consistently delivers outstanding segmentation results, confirming its superior performance in image segmentation tasks.

% \begin{figure*}[htbp]
%   \setlength{\tabcolsep}{2pt} 
%   \renewcommand{\arraystretch}{1}  
%   \adjustboxset{width=1.38in, height=1in, margin=0pt} 
%   \centering
%   \begin{tabular}{@{}cccc@{}} 
%     \adjustimage{}{Fig6-1.jpg} &
%     \adjustimage{}{Fig6-2.png} &
%     \adjustimage{}{Fig6-3.jpg} &
%     \adjustimage{}{Fig6-4.jpg} \\
    
%     \adjustimage{}{Fig6-5.jpg} &
%     \adjustimage{}{Fig6-6.png} &
%     \adjustimage{}{Fig6-7.jpg} &
%     \adjustimage{}{Fig6-8.jpg} \\
    
%     \adjustimage{}{Fig6-9.jpg} &
%     \adjustimage{}{Fig6-10.png} &
%     \adjustimage{}{Fig6-11.jpg} &
%     \adjustimage{}{Fig6-12.jpg} \\
    
%     \multicolumn{1}{c}{\footnotesize Image} & \multicolumn{1}{c}{\footnotesize SAM} & \multicolumn{1}{c}{\footnotesize FastSAM} & \multicolumn{1}{c}{\footnotesize FastSmoothSAM} \\
%   \end{tabular}
%   \caption{Comparison of segment everything results.}
%   \label{fig:exp_more}
% \end{figure*}

\subsubsection{Runtime comparisons}
This part provides a comparison of the runtime performance of SAM\cite{SAM}, FastSAM\cite{FastSAM}, MobileSAM\cite{MobileSAM, MobileSAMv2}, UnSAM\cite{UnSAM} and FastSmoothSAM on the COCO dataset\cite{coco}, the other recent competitors that cannot be configured and run in our machine (including EfficientSAM\cite{EfficientSAM} and SEEM\cite{SEEM}) are not compared. Table \ref{tab:runtime} compares various image sizes, highlighting the number of images processed, the average number of masks per image, and the average runtime per image and mask.

\begin{table}[htbp]
     \centering
     \renewcommand\arraystretch{1.3}
     \setlength\tabcolsep{3.2pt} %
     \caption{The runtime comparisons on the COCO dataset. FastSmoothSAM runs slower a little than FastSAM, while it outperforms other competitors. ('/' means not runnable on the corresponding GPU device.)}
    \begin{tabular}{lllllllll}
    \hline
    ~ & ~ & Image size & 333$\times$500 & 375$\times$500 & 427$\times$640 & 480$\times$640 & 612$\times$612 & 640$\times$640 \\ 
    \cmidrule{3-9}
    ~ & ~ & Number of image & 108 & 327 & 771 & 1386 & 100 & 53 \\
    \cmidrule{2-9} 
    ~ & \multirow{2}{*}{\begin{tabular}[c]{@{}l@{}}Average identified\\masks  per image\end{tabular}} & FastSAM & 43 & 39 & 48 & 50 & 40 & 40 \\
    ~ & ~ & {FastSmoothSAM} & {46} & {41} & {51} & {53} & {41} & {42} \\ 
    \hline
    \multirow{6}{*}{\begin{tabular}[c]{@{}l@{}}Nvidia\\ 
    GTX \\ 3060Ti\end{tabular}} & \multirow{2}{*}{\begin{tabular}[c]{@{}l@{}}Average runtime \\ per mask(ms)\end{tabular}} & FastSAM & 7.5 & 8.73 & 9.79 & 10.74 & 14.47 & 16.24 \\
    ~ & ~ & \uline{FastSmoothSAM} & \uline{9.71} & \uline{10.55} & \uline{11.63} & \uline{12.38} & \uline{16.11} & \uline{17.84} \\
    \cmidrule{2-9} 
    ~ & \multirow{5}{*}{\begin{tabular}[c]{@{}l@{}}Average runtime \\ per image (ms)\end{tabular}} & SAM & 10307 & 10612 & 13435 & 14858 & 14264 & 15795 \\
    ~ & ~ & MobileSAM & 2984 & 2991 & 3216 & 3171 & 3130 & 3146 \\
    ~ & ~ & UnSAM & / & / & / & / & / & / \\
    ~ & ~ & \textbf{FastSAM} & \textbf{325.77} & \textbf{343.74} & \textbf{465.51} & \textbf{537.55} & \textbf{576.41} & \textbf{649.23} \\
    ~ & ~ & \uline{FastSmoothSAM} & \uline{444.93} & \uline{436.92} & \uline{588.34} & \uline{651.2} & \uline{667} & \uline{745.4} \\ 
    \hline
    \multirow{7}{*}{\begin{tabular}[c]{@{}l@{}}Nvidia\\ GTX\\ 3090\end{tabular}} & \multirow{2}{*}{\begin{tabular}[c]{@{}l@{}}Average runtime \\ per mask (ms)\end{tabular}} & FastSAM & 5.6 & 6.52 & 7.4  & 7.98 & 10.6 & 11.44 \\
    ~ & ~ & \uline{FastSmoothSAM} & \uline{6.32} & \uline{6.92} & \uline{8.19} & \uline{8.65} & \uline{11.3} & \uline{11.99} \\
    \cmidrule{2-9} 
    ~ & \multirow{5}{*}{\begin{tabular}[c]{@{}l@{}}Average runtime\\ per image (ms)\end{tabular}} & SAM & 6278 & 6468 & 8525 & 9504 & 9063 & 10302 \\
    ~ & ~ & MobileSAM & 2191 & 2149 & 2354 & 2333 & 2225 & 2291 \\
    ~ & ~ & UnSAM & 3320 & 3189 & 4554 & 4874 & 5287 & 5655 \\
    ~ & ~ & \textbf{FastSAM} & \textbf{243.12} & \textbf{256.12} & \textbf{352.02} & \textbf{398.71} & \textbf{422.03} & \textbf{512.25} \\
    % \rowcolor{gray!20}
    ~ & ~ & \uline{FastSmoothSAM} & \uline{289.69} & \uline{285.98} & \uline{414.24} & \uline{453.85} & \uline{467.87} & \uline{559.14} \\ 
    \hline
    \end{tabular}
    \label{tab:runtime}
\end{table}

It is evident that our method generally exhibits a slightly higher average runtime (about $2ms$ per mask in Nvidia GTX 3060Ti and $0.7ms$ per mask in Nvidia GTX 3090) compared to FastSAM, but remains significantly faster than others. These comparisons illuminate the trade-offs between the quality of masks and the computational efficiency offered by each method. Due to these characteristics, our method is particularly suitable for scenarios where only a few primary objects need to be extracted from an image.

\begin{figure}[htbp]
  \centering
  \includegraphics[width=5.8in]{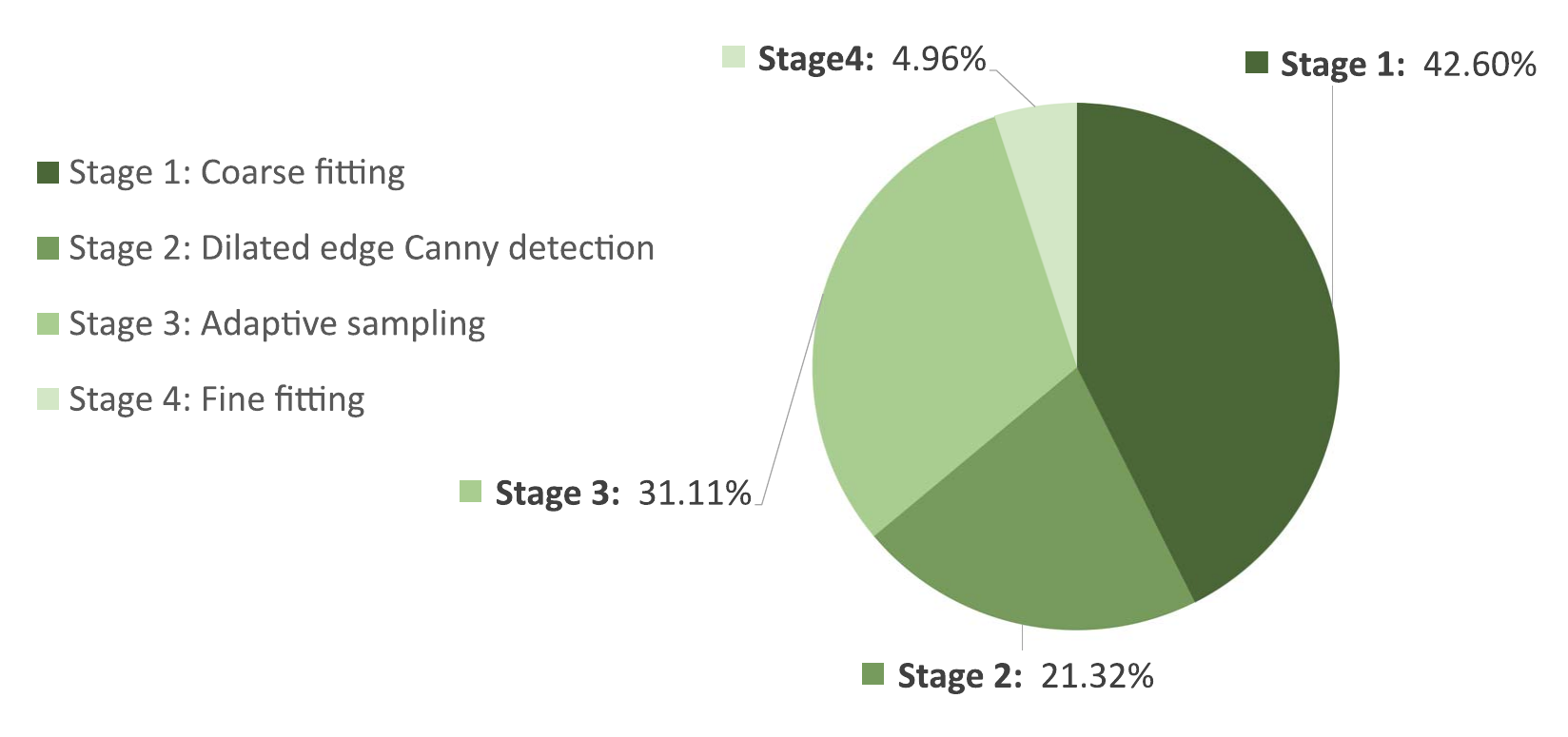}
  \caption{The additional time overhead distribution of FastSmoothSAM}
  \label{fig:additonal_time_cost}
\end{figure}

We also accounted for the extra time overhead, and the distribution of additional time costs for our algorithm is shown in Fig. \ref{fig:additonal_time_cost}, which reveals that the main costs are incurred during Stage 1 and Stage 3, which account for $42.60\%$ and $31.11\%$ of the total time, respectively. Because extracting points in Stage 1 takes too long.

We compared the runtime memory consumption of three models—SAM, FastSAM, and our FastSmoothSAM—using 5,000 images from the COCO dataset, as shown in the Table \ref{tab:Memory_consumption}. The results indicate that SAM has the highest memory usage among the three due to its use of ViT as the backbone. Although ViT provides powerful feature extraction capabilities, its high computational complexity results in significant memory demands. FastSAM reduces memory consumption by leveraging YOLOv8 for optimization, simplifying the model complexity and improving runtime efficiency. Building on this foundation, our FastSmoothSAM further optimizes performance by introducing curve-fitting-based smoothing, which effectively reduces runtime memory fluctuation. Overall, FastSmoothSAM maintains runtime efficiency and accuracy while achieving the lowest memory consumption, demonstrating significant advantages and broad potential for practical applications.

\begin{table}[htbp]
    \renewcommand\arraystretch{1.3}
    \centering
    \caption{Comparison of Memory Consumption Among SAM, FastSAM, and FastSmoothSAM on COCO Dataset.}
    \setlength{\tabcolsep}{12mm}
    \begin{tabular}{ccc} 
    \hline
     Model & \begin{tabular}[c]{@{}c@{}} Average Memory \\ Consumption (\%) \end{tabular} & \begin{tabular}[c]{@{}c@{}} Maximum Memory \\ Consumption (\%) \end{tabular} \\ 
    \hline
    SAM & 13.09 & 19.53 \\
    FastSAM & 8.06 & 8.68 \\
    FastSmoothSAM & \textbf{7.99} & \textbf{8.36} \\
    \hline
    \end{tabular} 
    \label{tab:Memory_consumption}
\end{table}

\subsubsection{Evaluation by metrics}
To quantitatively compare FastSAM, FastSmoothSAM(Ours) and U-Net\cite{UNet}, we conducted experiments on the COCO dataset\cite{coco}, UNet dataset\footnote{\url{https://github.com/milesial/Pytorch-UNet}} and the BSDS500 dataset\footnote{\url{https://github.com/BIDS/BSDS500}}\cite{BSDS500}. We compared the results using three metrics: curvature, Fréchet distance (FD), and edge detection, as detailed below.

1) Curvature is commonly used to evaluate the smoothness performance of curves. Measures the rate of change at specific points, which helps quantify the steepness or extent of change along the curve. By calculating the average rate of curvature change between adjacent points, an overall smoothness index can be determined. This method is widely applied in various fields. Evaluating the smoothness of curves improves our understanding of the characteristics and trends in the data, providing a basis for further analysis and application. The formula used is as follows:

\begin{equation}
\label{equ:curvature}
\begin{aligned}
\kappa_{i} = \frac {x_{i-1}(y_i-y_{i+1}) - x_i(y_{i-1}-y_{i+1}) + x_{i+1}(y_{i-1}-y_i)}{(x_{i-1}-x_i)^2 + (y_{i-1}-y_i)^{3/2}}.
\end{aligned}
\end{equation}

This formula calculates the change in curvature at each point, considering the two points before and after it. Here, $\kappa_{i}$ represents the curvature at the $i$-th point. This measurement is used to assess the curvature of a given set of points.

2) The Fréchet distance (FD) is a mathematical concept used to measure the similarity between two curves. It involves selecting a parameter on each curve and considering the shortest distance between the corresponding points on these curves. Then, among all possible choices of parameters, the maximum of these shortest distances is determined. This maximum shortest distance represents the Fréchet distance, which reflects the greatest degree of difference between the two curves. The formula used is as follows:

\begin{equation}
F(P, Q)=\inf_{\alpha,\beta}\max_{t\in [0, 1]} \{dist (P(\alpha (t)), Q(\beta (t))) \},
\end{equation}
where $P(\alpha (t))$ and $Q(\beta (t))$ represent the corresponding points on curves $P$ and $Q$ respectively.

The results in Table \ref{tab:metirc_comparisons} indicate that compared to FastSAM, U-Net in COCO, UNet dataset, FastSmoothSAM exhibits better smoothness in contour edges. Specifically, our findings demonstrate that the contour edges produced by our method are relatively smooth, while those from FastSAM often suffer from scale loss and show jagged edges. Additionally, in terms of the computed Fréchet distance, our method achieves superior results, with a smaller F-dist indicating better performance.

\begin{table}[htbp]
    \renewcommand\arraystretch{1.3}
    \centering
    \caption{The evaluation metrics for the differences masks obtained from between the FastSAM, FastSmnoothSAM, U-Net and ground truth values.}
    \setlength{\tabcolsep}{5.4mm}
    \begin{tabular}{cccccc} 
    \hline
     Dataset & Model & \begin{tabular}[c]{@{}c@{}} Number \\ of Image \end{tabular} & \begin{tabular}[c]{@{}c@{}} Mean \\ Curvature \end{tabular}  & \begin{tabular}[c]{@{}c@{}} Curvature \\ Variance \end{tabular}  & FD\\
    \hline
    \multirow{2}{*}{\begin{tabular}[c]{@{}l@{}}COCO \end{tabular}} & FastSAM & 80 & .064 & .0065 & 2.24 \\
    ~ & FastSmoothSAM & 80 & \textbf{.055} & \textbf{.0042} & \textbf{1.81} \\
    \hline
    \multirow{3}{*}{\begin{tabular}[c]{@{}l@{}}UNet \end{tabular}} & U-Net & 200 & .363 & .2572 & 172.68 \\ ~ & FastSAM & 200 & .088 & .0128 & 107.22 \\
    ~ & FastSmoothSAM & 200 & \textbf{.080} & \textbf{.0107} & \textbf{104.49} \\
    \hline
    \end{tabular} 
    \label{tab:metirc_comparisons}
\end{table}

\begin{figure}[htbp]
  \centering
  \subfigure{\includegraphics[width=1.38in]{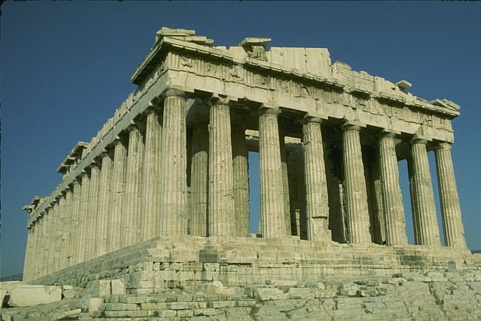}}
  \subfigure{\includegraphics[width=1.38in]{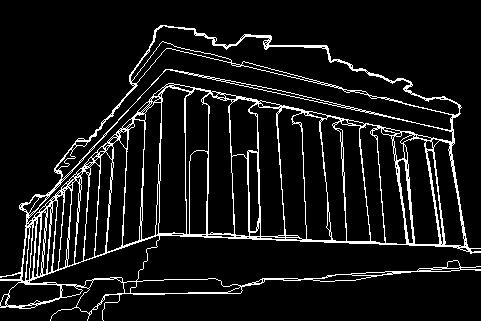}}
  \subfigure{\includegraphics[width=1.38in]{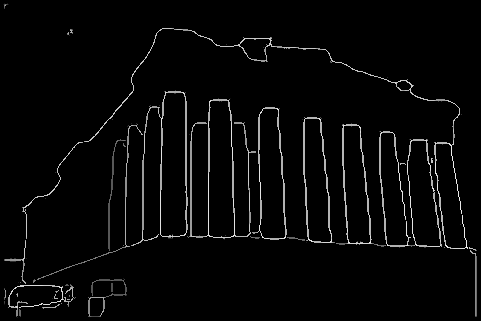}}
  \subfigure{\includegraphics[width=1.38in]{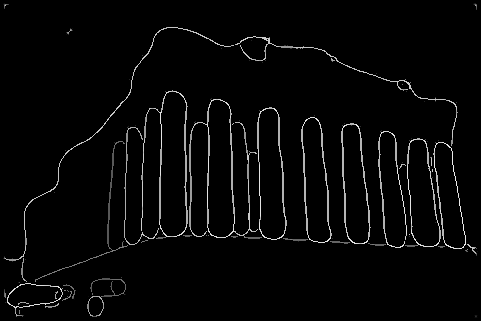}}
  \subfigure{\includegraphics[width=1.38in]{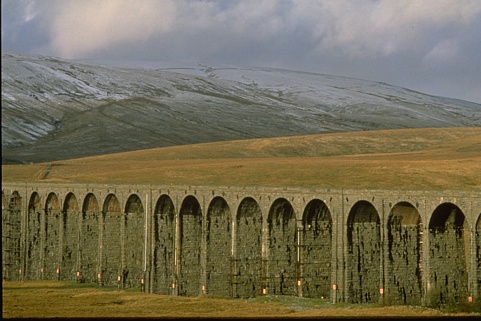}}
  \subfigure{\includegraphics[width=1.38in]{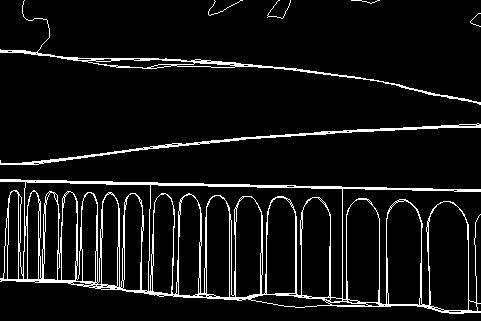}}
  \subfigure{\includegraphics[width=1.38in]{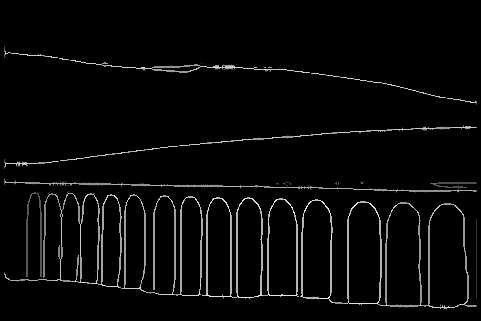}}
  \subfigure{\includegraphics[width=1.38in]{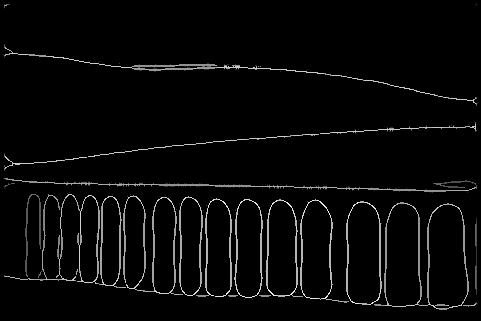}}
  \\\leftline{\scriptsize \hspace*{0.5in}Image \hspace{0.75in} Ground truth \hspace{0.7in} FastSAM \hspace{0.6in} FastSmoothSAM}
  \centering
  \caption{Zero-shot edge prediction on BSDS500. The bad case of  FastSmoothSAM using B-Spline for fitting objects with clear sharp edges}
  %achieves comparable results (bottom). However, the top shows that FastSmoothSAM using B-Spline for fitting is prone to angle distortion.}
  \label{fig:eval_detection}
\end{figure}

\begin{figure}[htbp]
  \centering
  \subfigure[Small fruits]{\includegraphics[width=2.7in, height=1.8in]{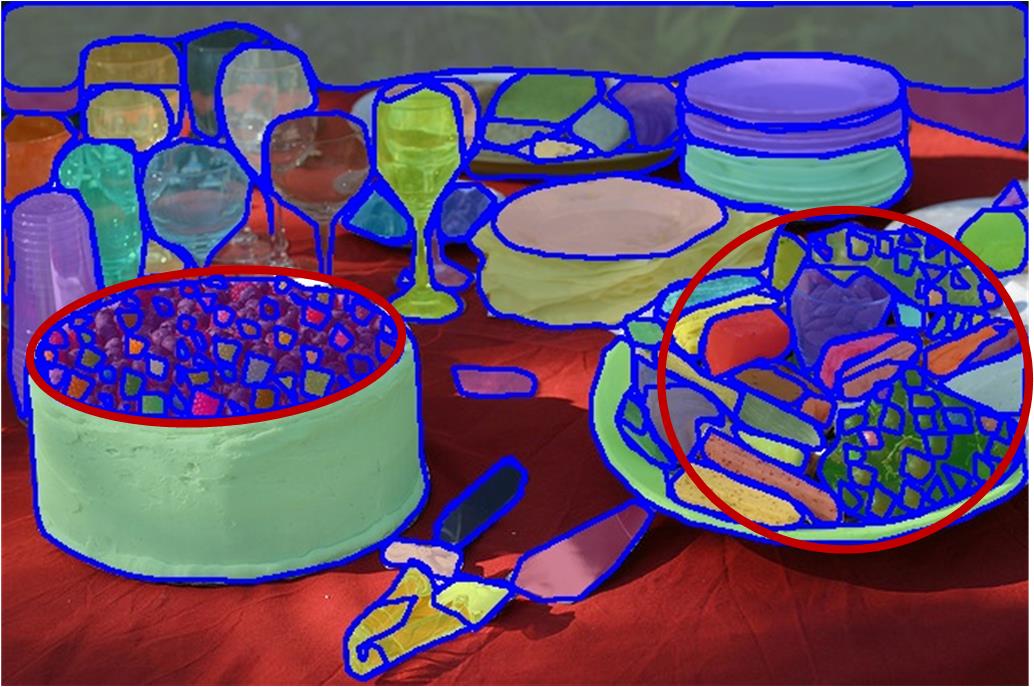}}
  \hspace{4mm}
  \subfigure[Buttons of keyboard]{\includegraphics[width=2.7in, height=1.8in]{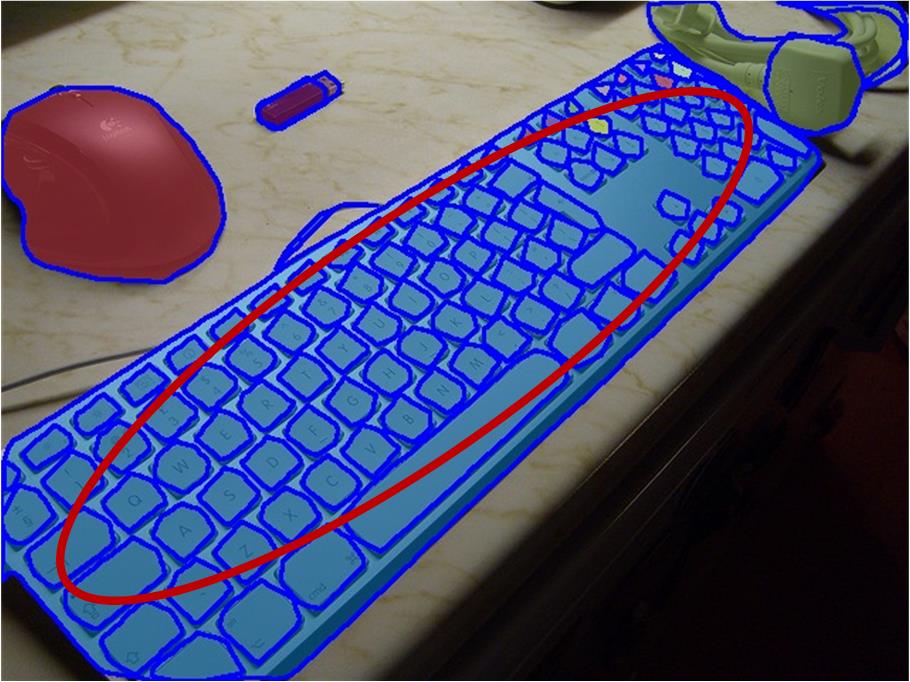}}
  % \subfigure[Pointed pentagram]{\includegraphics[width=2.2in]{bad/bad4.jpg}}\\
  \centering
  % ~(a) small fruits~~~~~~~~~~~~~~~~~(b) buttons of keyboard 
  \caption{Some bad examples of FastSmoothSAM. In handling small-sized segmentation masks, FastSmoothSAM may yield poorly fitted contours. For instance, in (a), numerous small ball-like fruits within the red ellipse are incorrectly identified which are far from their actual shapes. Similarly, in (b), although the keyboard buttons are square or rectangular, the recognized shapes are almost skewed}
  \label{fig:bad_case}
\end{figure}

3) FastSmoothSAM is evaluated for basic low-level texture smoothing tasks using the BSDS500 dataset. We use a simplified version of the automatic smoothing mask generation pipeline. After the network administrator remove redundant masks, the smoothed texture map is computed using Gaussian filtering\cite{Gaussian} without thresholding the smoothed probability map. This process is concluded with lightweight post-processing, including a texture NMS\cite{Canny} step. The experimental results are shown in Table \ref{tab:eval_detection}. It can be seen that our obtained AP and R50 are very close to those of FastSAM, but slightly lower. The main reasons are: 

a. In the BSDS500 dataset, FastSAM has few jagged edges. 

b. There are many objects with clear sharp edges as illustrated in Fig. \ref{fig:eval_detection}. Due to the inherent properties of B-Spline curves, the smooth nature of B-Spline makes it does not perform well when fitting sharp-edged objects. Additionally, for small objects, the limited number of edge sample points can lead to significant errors in the fitting process, as Fig. \ref{fig:bad_case} shows, small objects are all distorted. 

\begin{table}[htbp]
    \renewcommand\arraystretch{1.3}
    \centering
    \caption{Zero-shot transfer to edge detection on BSDS500. Evaluation Data of other methods is from \cite{SAM}.}
    \setlength{\tabcolsep}{7.5mm}
    \begin{tabular}{lc|cccc} 
    \hline
    Method & Year & ODS & OIS & AP & R50\\
    \hline
    HED\cite{HED} & 2015 & .778 & .808 & .840 & .923\\
    EDETR\cite{EDETR} & 2022 & .840 & .858 & .896 & .930 \\
    \multicolumn{6}{l}{$zero$-$shot$ $transfer$ $methods:$}\\
    Sobel filter & 1968 & .539 & - & - & - \\
    Canny\cite{Canny} & 1986 & .600 & .640 & .580 & - \\
    Felz-Hutt\cite{Felz-Hutt} & 2004 & .610 & .640 & .560 & - \\
    SAM\cite{SAM} & 2023 & .768 & .786 & .794 & .928 \\
    FastSAM\cite{FastSAM} & 2023 & .734 & .753 & .634 & .696 \\
    FastSmoothSAM & 2024 & .737 & .749 & .620 & .678 \\
    \hline
    \end{tabular}
    \label{tab:eval_detection}
\end{table}

\subsection{Ablation Study}
\begin{figure}[htbp]
  \centering
  \includegraphics[width=5.5in]{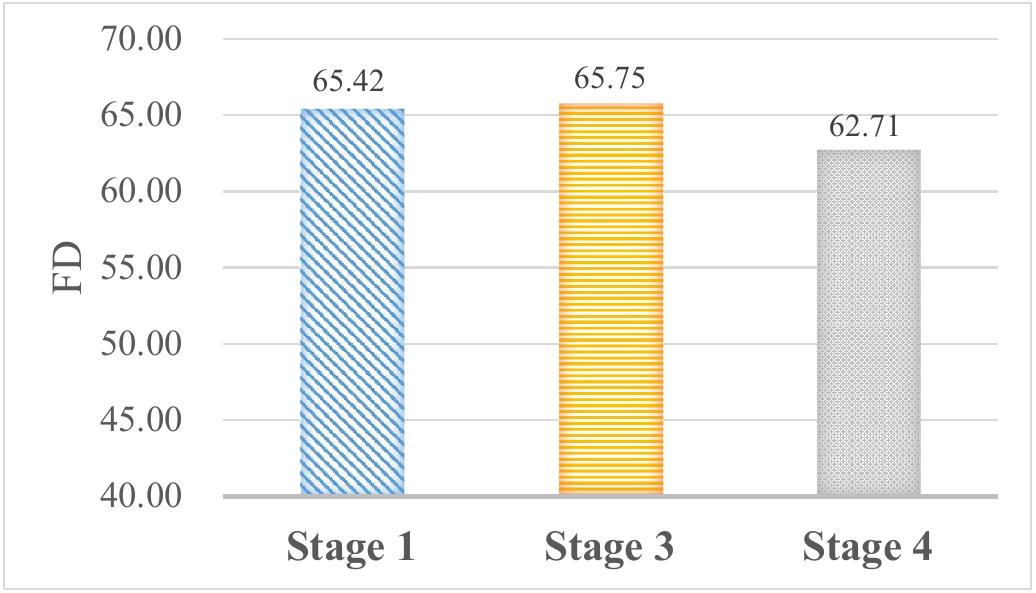}
  \caption{Performance Comparison of FastSmoothSAM at Different Stages (FD Metric) in PASCAL VOC dataset. Stage 2 is not included in the comparison scope because the disordered edge points cannot form a continuous closed curve}
  \label{fig:Ablation}
\end{figure}

In this section, we analyze the performance of the FastSmoothSAM model at different stages in the PASCAL VOC dataset\footnote{\url{http://host.robots.ox.ac.uk/pascal/VOC/voc2012/}}\cite{pascal}, using the FD metric. The experimental results demonstrate that models that only complete Stage 1 or the first three stages (without Stage 4) perform significantly worse than those executing all four stages. As shown in Fig. \ref{fig:Ablation}, a lower FD value indicates better alignment with the original data. Executing only stage one leads to curve overfitting, making it difficult to accurately match the true edge contours; Models lacking stage four, which involves curve fitting, produce edges with insufficient smoothness. Additionally, since Stage 2 generates disordered edge points after dilation and cannot form continuous closed curves, it is excluded from quantitative comparison. These findings validate the importance of completing all four stages to accurately reconstruct the geometric structure of the original edges.

\section{Conclusion and future work}
Fast Segment Anything (FastSAM) addresses the limitations of high memory consumption and prolonged inference times found in the Segment Anything Model (SAM), but often produces jagged edges that deviate from the true contours of objects. 

To overcome this challenge, the application of the B-Spline curve has proven effective in enhancing edge recognition quality. By leveraging the shape control and geometric construction capabilities of B-Spline, we implemented a four-stage refining process that utilizes two rounds of B-Spline curve fitting techniques. This approach effectively smooths jagged edges, ensuring that the segmentation more closely aligns with the actual shapes of objects. The refined method significantly improves the visual quality and analytical accuracy of object edges while preserving essential geometric information.

This advancement not only optimizes the performance and utility of SAM and FastSAM but also contributes to more accurate and visually better segmentation outcomes in real-time applications.

However, due to the inherent properties of the B-spline curve, they do not perform well when fitting sharp-edged objects, and for small objects, the limited number of edge sample points can lead to significant errors in the fitting process, as shown in Fig. \ref{fig:bad_case}. Our future work will focus on improving the accuracy of fitting sharp-edged and small objects, aiming to develop more robust algorithms capable of adapting to the specific characteristics of the objects.

\section*{Acknowledgment}
This work was supported by the National Natural Science Foundation of China (No.61673186); the Natural Science Foundation of Fujian Province,  China (No.2021J01317,  2020J05059); and Fujian Provincial Natural Science Foundation General Project (No.2022J01316).

\section*{Statements and Declarations}
We declare that we have no known competing financial interests or personal relationships that could have appeared to influence the work reported in this paper.

\bibliography{main}% common bib file

\end{document}